\documentclass{article}

\usepackage[final]{corl_2018}
\usepackage{amsmath}
\usepackage{amssymb}
\usepackage{graphics}
\usepackage{epsfig}
\usepackage{mathptmx}
\usepackage{nicefrac}
\usepackage{mathtools}
\usepackage{times}
\usepackage[font={small,it}]{caption}
\usepackage{hyperref}
\usepackage{wrapfig}
\usepackage{authblk}
\usepackage{titlesec}
\usepackage[caption=false,margin=0pt]{subfig}
\usepackage{algorithm}
\usepackage{algorithmic}
\usepackage{svg}
\setcitestyle{square}
\usepackage{authblk}
\titleformat{\subsection}[runin]
{\normalfont\bfseries}{\thesubsection}{1em}{}

\title{Grasp2Vec: Learning Object Representations\\ from Self-Supervised Grasping}

\author[*,1]{Eric Jang\hspace{0.3em}}
\author[*,2,$\dagger$]{Coline Devin\hspace{0.4em}}
\author[1]{Vincent Vanhoucke}
\author[1,2]{Sergey Levine}
\affil[*]{\footnotesize Both authors contributed equally}
\affil[1]{\footnotesize Google Brain}
\affil[2]{\footnotesize Department of Electrical Engineering and Computer Sciences, UC Berkeley}
\affil[$\dagger$]{\footnotesize Work done while author was interning at Google Brain}
\affil[ ]{\texttt{\{ejang, vanhoucke, slevine\}@google.com}}
\affil[]{\texttt{coline@eecs.berkeley.edu}}
\begin{document}
\maketitle

\thispagestyle{empty}

\newcommand{\cmt}[1]{{\footnotesize\textcolor{red}{#1}}}
\newcommand{\note}[1]{\cmt{Note: #1}}
\newcommand{\todo}[1]{\cmt{TODO: #1}}

\newcommand{\scenepre}{s_\text{pre}} 
\newcommand{\scenepost}{s_\text{post}} 
\newcommand{\outcome}{\mathbf{o}} 
\newcommand{\goal}{\mathbf{g}} 

\newcommand{\phimap}{\phi_{s,\text{spatial}}} 
\newcommand{\phimapg}{\phi_{g,\text{spatial}}} 

\newcommand{\phis}{\phi_s}
\newcommand{\phig}{\phi_o}

\newcommand{\bs}{\mathbf{s}}
\newcommand{\ba}{\mathbf{a}}
\newcommand{\reward}{\mathbf{r}}
\newcommand{\rewardany}{\reward_\text{any}}
\newcommand{\bsp}{\mathbf{s^\prime}} 
\newcommand{\othergoal}{\mathbf{g^\prime}}

\newcommand{\piany}{\pi_\text{any}}
\newcommand{\Qpi}{Q_{\pi}} 

\newcommand{\replaybuffer}{B}

\newcommand{\E}[1]{\mathbb{E}\left[#1\right]} 
\newcommand{\norm}[1]{\lVert \mathbf{#1} \rVert}

\newcommand{\numkukareal}{\todo{150k}}
\newcommand{\numkukasim}{{500,000}}

\begin{abstract}
Well structured visual representations can make robot learning faster and can improve generalization. In this paper, we study how we can acquire effective object-centric representations for robotic manipulation tasks without human labeling by using autonomous robot interaction with the environment. Such representation learning methods can benefit from continuous refinement of the representation as the robot collects more experience, allowing them to scale effectively without human intervention.
Our representation learning approach is based on object persistence: when a robot removes an object from a scene, the representation of that scene should change according to the features of the object that was removed. We formulate an arithmetic relationship between feature vectors from this observation, and use it to learn a representation of scenes and objects that can then be used to identify object instances, localize them in the scene, and perform goal-directed grasping tasks where the robot must retrieve commanded objects from a bin. The same grasping procedure can also be used to automatically collect training data for our method, by recording images of scenes, grasping and removing an object, and recording the outcome. Our experiments demonstrate that this self-supervised approach for tasked grasping substantially outperforms direct reinforcement learning from images and prior representation learning methods.
\end{abstract}
\keywords{instance grasping, unsupervised learning, reinforcement learning}

\section{Introduction}
Robotic learning algorithms based on reinforcement, self-supervision, and imitation can acquire end-to-end controllers from  images for diverse tasks such as robotic mobility~\cite{ross2013learning,bojarski2016end} and object manipulation~\cite{ghadirzadeh2017deep,levine2016learning}. These end-to-end controllers acquire perception systems that are tailored to the task, picking up on the cues that are most useful for the control problem at hand. However, if our aim is to learn generalizable robotic skills and endow robots with broad behavior repertoires, we might prefer perceptual representations that are more structured, effectively disentangling the factors of variation that underlie real-world scenes, such as the persistence of objects and their identities.  A major challenge for such representation learning methods is to retain the benefit of self-supervision, which allows leveraging large amounts of experience collected autonomously, while still acquiring the structure that can enable superior generalization and interpretability for downstream tasks.

In this paper, we study a specific instance of this problem: acquiring object-centric representations through autonomous robotic interaction with the environment. By interacting with the real world, an agent can learn about the interplay of perception and action.
For example, looking at and picking up objects enables a robot to discover relationships between physical entities and their surrounding contexts. If a robot grasps something in its environment and lifts it out of the way, then it could conclude that anything still visible was not part of what it grasped. It can also look at its gripper and see the object from a new angle. Through active interaction, a robot could learn which pixels in an image are graspable objects and recognize particular objects across different poses without any human supervision.

While object-centric representations can be learned from semantically annotated data (e.g., the MSCOCO dataset~\cite{lin2014microsoft}), this precludes continuous self-improvement: additional experience that the robot collects, which lacks human annotations, is not directly used to improve the quality and robustness of the representation. In order to improve automatically, the representation must be self-supervised. In that regime, every interaction that the robot carries out with the world improves its representation.

\begin{figure}[t]
\centering
\vspace{-0.05in}
\includegraphics[width=\textwidth]{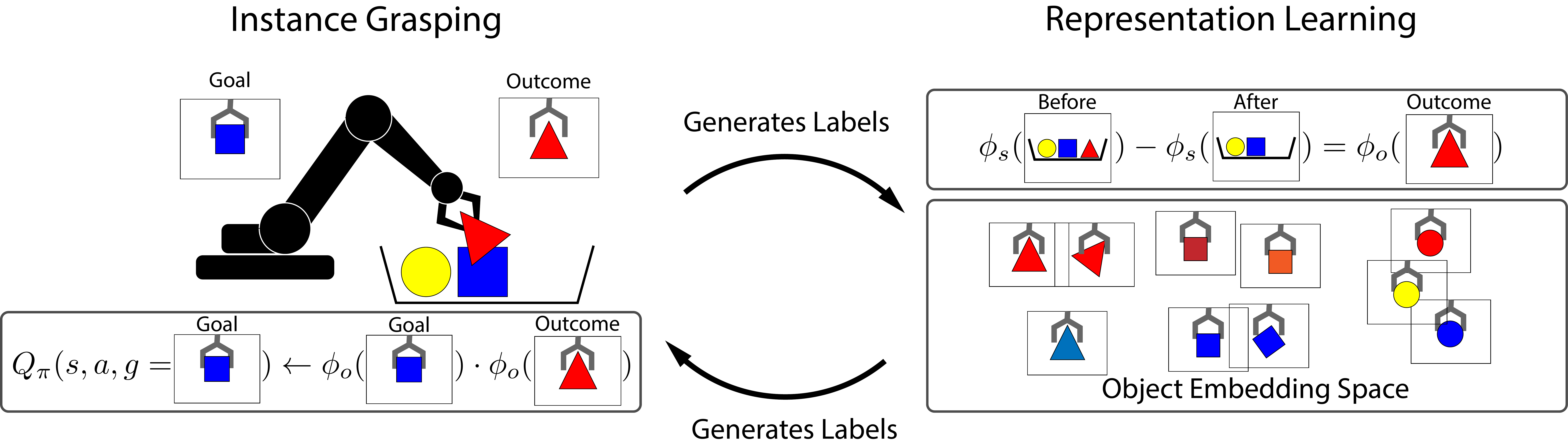}
\caption{Instance grasping and representation learning processes generate each other's labels in a fully self-supervised manner. \textbf{Representation learning from grasping:} A robot arm removes an object from the scene, and observes the resulting scene and the object in the gripper. We enforce that the difference of scene embeddings matches the object embedding. \textbf{Supervising grasping with learned representations:} We use a similarity metric between object embeddings as a reward for instance grasping, removing the need to manually label grasp outcomes.}

\label{fig:method_schematic}
\vspace{-0.2in}
\end{figure}

Our representation learning method is based on object persistence: when a robot picks up an object and removes it from the scene, the representation of the scene should change in a predictable way. We can use this observation to formulate a simple condition that an object-centric representation should satisfy: the features corresponding to a scene should be approximately equal to the feature values for the same scene after an object has been removed, minus the feature value for that object (see Figure~\ref{fig:method_schematic}). We train a convolutional neural network feature extractor based on this condition, and show that it can effectively capture individual object identity and encode sets of objects in a scene without any human supervision.

Leveraging this representation, we propose learning a self-supervised grasping policy conditioned on an object feature vector or image.
While labeling whether the correct object was grasped would typically require human supervision, we show that the
similarity between object embeddings (learned with our method) provides an equally good reward signal.

Our main contribution is grasp2vec, an object-centric visual embedding learned with self-supervision. We demonstrate how this representation can be used for object localization, instance detection, and goal-conditioned grasping, where autonomously collected grasping experience can be relabeled with grasping goals based on our representation and used to train a policy to grasp user-specified objects. We find our method outperforms alternative unsupervised methods  in a simulated goal-conditioned grasping results benchmark. Supplementary illustrations and videos are at \url{https://sites.google.com/site/grasp2vec/home}
\section{Related Work}

\paragraph{Unsupervised representation learning.}
Past works on interactive learning have used egomotion of a mobile agent or poking motions \cite{agrawal2016learning, ebert2017self, pinto2016curious, jayaraman2015learning, jonschkowski2015learning}
to provide data-efficient learning of perception and control. Our approach learns representations that abstract away position and appearance, while preserving object identity and the combinatorial structure in the world (i.e., which objects are present) via a single feature vector. Past work has also found that deep representations can exhibit intuitive linear relationships, such as in word embeddings~\cite{mikolov2013distributed}, and in face attributes~\cite{radford2015unsupervised}. Wang et. al represent actions as the transformation from precondition to effect in the action recognition domain~\cite{wang2016actions}. While our work shares the idea of arithmetic consistency over the course of an action, we optimize a different criterion and apply the model to learning policies rather than action recognition.

\vspace{-0.1in}
\paragraph{Self-supervised grasping.}

A recent body of work has focused on deep visuomotor policies for picking up arbitrary objects from RGB images \cite{pinto2016supersizing, levine2016learning, lenz2015deep,zeng2017multi}.
By automatically detecting whether some object was grasped, these methods can learn without human supervision.
Ten Pas et al. combine object detection with grasping to be able to grasp target objects by providing class labeled training data for each object~\cite{ten2017grasp} and using grasp predictions as object proposals.
\begin{wrapfigure}{r}{0.4\textwidth}
\vspace{-0.05in}
\begin{center}
\includegraphics[width=.4\textwidth]{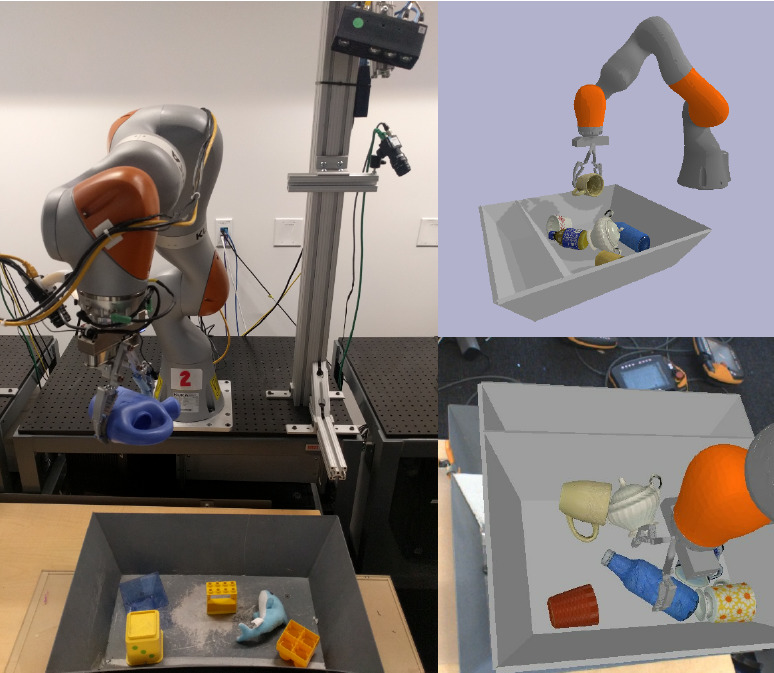}
\end{center}
\caption{Self-supervised robotic setup for learning grasp2vec and goal-conditioned grasping. Left: A KUKA iiwa uses a monocular RGB camera to pick up objects from a bin. Right: The same setup in simulation. Bottom right: Example images that are fed into the Q function shown in Figure~\ref{fig:grasp2vec}.}
\label{fig:kuka_setup}
\vspace{-0.2in}
\end{wrapfigure}

Addressing the task of instance grasping requires inferring whether the correct object was grasped. Jang et al. propose a system where the robot presents images of objects to the camera after it has grasped them and attempts to label their class given human labels.~\cite{jang2017semantic}
Fang et al. obtain labels in simulation and use domain adaptation to generalize the policy to real-world scenes, which requires solving a simulation-to-reality transfer problem~\cite{fang2017multi}.

 In the standard RL setting, several past works have studied labeling off-policy behavior with ``unintentional rewards''~\cite{hindsight2017,cabi2017intentional}.
However, such algorithms do not address how to detect whether the desired goal was achieved, which is non-trivial outside of the simulator. Our methods circumvent the problem of labeling entirely via self-supervised representation learning. To our knowledge, this is the first work that learns the instance grasping task in a completely label-free manner.

Concurrent work by Florence et al.~\cite{florence2018dense} uses a pixelwise
contrastive loss change detection in depth images of grasped objects to learn descriptors.
 In contrast to this work, we report quantitative results that indicate our method achieves high accuracy in identifying, localizing, and grasping objects, even when the instance is specified with a different view, under heavy occlusion, or under deformation.

\section{Grasp2Vec: Representation Learning from Grasping}
\label{sec:grasp2vec_method}

Our goal is to learn an object-centric embedding of images. The embeddings should represent objects via feature vectors, such that images with the same objects are close together, and those with different objects are far apart. Because labels indicating which objects are in an image are not available, we rely on a self-supervised objective. Specifically, we make use of the fact that, when a robot interacts with a scene to grasp an object, this interaction is quantized: it either picks up one or more whole objects, or nothing. When an object is picked up, we learn that the initial scene must have contained the object, and that the scene after must contain one fewer of that object. We use this concept to structure image embeddings by asking that feature difference of the scene before and after grasping is close to the representation of the grasped object.

We record grasping episodes as images triples:
$(\scenepre, \scenepost, \outcome)$,
where $\scenepre$ is an image of the scene before grasping, $\scenepost$ is the same scene after grasping, and $\outcome$ is an image of the grasped object held up to the camera. The specific grasping setup that we use, for both simulated and real image experiments, is described in Section~\ref{sec:experiments}.
Let $\phis(\scenepre)$ be a vector embedding of the input scene image (i.e., a picture of a bin that the robot might be grasping from).
Let $\phig(\outcome)$ be a vector embedding of the outcome image,
such that $\phis(\scenepre)$ and $\phig(\outcome)$ are the same dimensionality.
We can express the logic in the previous paragraph as an arithmetic constraint on these vectors: we would like
$(\phis(\scenepre)-\phis(\scenepost))$ to be equal to $\phig(\outcome)$. We also would like the embedding to be non-trivial, such that $(\phis(\scenepre)-\phis(\scenepost))$ is far from the embeddings of other objects that were not grasped.

\begin{figure}
\vspace{-0.05in}
\includegraphics[width=\textwidth]{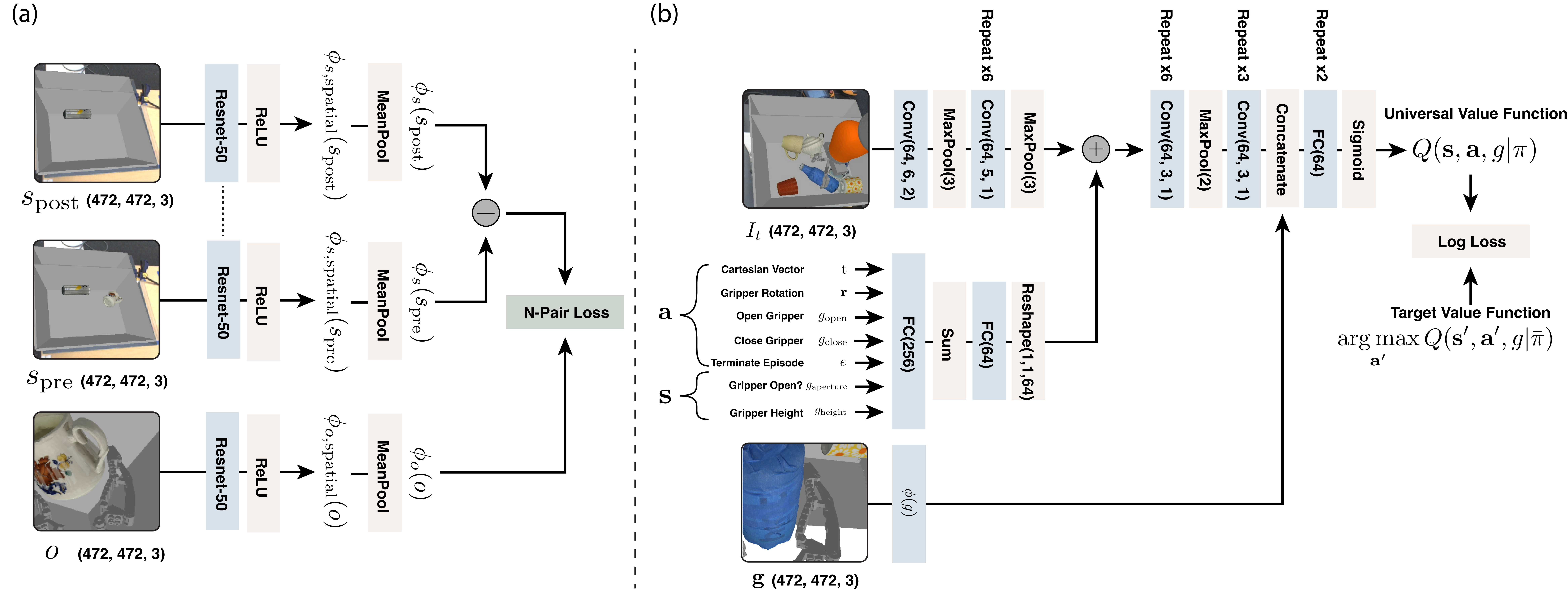}
\caption{(a) The Grasp2Vec architecture. We use the first 3 blocks of Resnet-50 V2 to form $\phis$ and $\phig$, which are randomly initialized. $\phis(\scenepre)$ and $\phis(\scenepost)$ have tied weights. The output of the third resnet block is passed through a ReLu to yield a spatial feature map we call $\phi_\text{spatial}$. This is then mean pooled globally to output the single vector embddings $\phis$ and $\phig$. The n-pairs loss is applied as described in Section~\ref{sec:npairs}. (b) The instance grasping architecture is conditioned on Grasp2Vec goal embeddings of goal images, and is trained via Q-learning.}
\label{fig:grasp2vec}
\vspace{-0.2in}
\end{figure}
\vspace{-0.1in}
\paragraph{Architecture.}
In order to embed images of scenes and outcomes, we employ convolutional neural networks to represent both $\phis$ and $\phig$. The two networks are based on the ResNet-50~\cite{he2016deep} architecture followed by a ReLU (see Figure~\ref{fig:grasp2vec}), and both produce 3D ($H$x$W$x1024) convolutional feature maps  $\phimap$ and $\phimapg$. Since our goal is to obtain a single vector represention of objects and sets of objects, we convert these maps into feature vectors by globally average-pooling: $\phis = \nicefrac{\sum_{i<H}\sum_{j<W}\phimap(X)[i][j]}{H*W}$ and equivalently for $\phig$. The motivation for this architecture is that it allows $\phimap$ to encode object position by which receptive fields produce which features. By applying a ReLU non-linearity to the spatial feature vectors, we constrain object representations to be non-negative. This ensures that a set of objects can only grow as more objects are added; one object cannot be the inverse of another.

\vspace{-0.1in}
\paragraph{Objective.}
\label{sec:npairs}
The problem is formalized as metric learning, where the desired metric places $(\phis(\scenepre)-\phis(\scenepost))$ close to $\phig(\outcome)$ and far from other embeddings. Many metric learning losses use the concept of an ``anchor" embedding and a ``positive" embedding, where the positive is brought closer to the anchor and farther from the other ``negative" embeddings.
One way to optimize this kind of objective is to use the n-pairs loss~\cite{Sohn2016Npairs} to train the encoders $\phis$ and $\phig$,
such that paired examples (i.e., $(\phis(\scenepre)-\phis(\scenepost))$ and $\phig(\outcome)$
are pushed together, and unpaired examples are pushed apart.
Rather than processing explicit (anchor, positive, negative) triplets, the n-pairs loss treats all \textit{other} positives in a minibatch as negatives for an (anchor, positive) pair. Let $i$ index into the anchors $a$ of a minibatch and let $j$ index into the positives $p$. The objective is to maximize $a_i\top p_i$ while minimizing $a_i\top p_{j\neq i}$.
The loss is the the sum of softmax cross-entropy losses for each anchor $i$ accross all positives $p$.
\[\text{NPairs}(a, p) = \sum_{i<B} -\log \left( \frac{e^{a_i\top p_i}}{\sum_{j<B} e^{a_i,p_j}} \right) +\lambda( ||a_i||_2^2+||p_i||_2^2).\]
The hyperparameter $\lambda$ regularizes the embdding magnitudes and $B$ is the batch size. In our experiments, $\lambda=0.0005$ and $B=16$.
This loss is asymmetric for anchors and positives, so we evaluate with the embeddings in both orders, such that our final training objective is:
\[\mathcal{L}_\text{Grasp2Vec} = \text{NPairs}((\phis(\scenepre)-\phis(\scenepost)), \phig(\outcome)) + \text{NPairs}( \phig(\outcome), (\phis(\scenepre)-\phis(\scenepost))).\]

\begin{figure*}
\vspace{-0.05in}
\centering
\subfloat[Nearest neighbors of goal and scene images.\label{fig:neighbors}]{
\includegraphics[width=\textwidth]{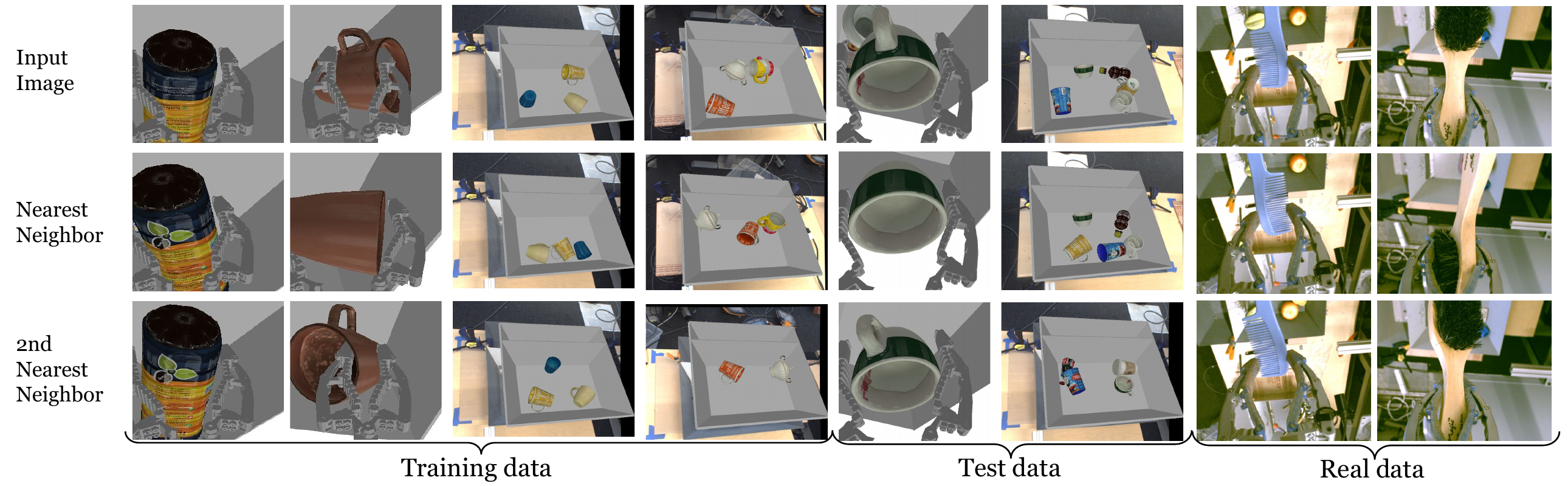}}
\\
\subfloat[Localization results. The heatmap is defined as $\phig(\outcome)\top \phimap(\scenepre)$, resulting in a $H\times W \times 1$ array. Note that the embeddings were never trained on this task, nor was any detection supervision provided. The fourth column shows a failure case, where two different mugs have similar features and the argmax is on the wrong mug. The right half shows results on real images. The representations trained on real grasping results are able to localize objects with high success. \label{fig:detection}]{
\includegraphics[width=\textwidth]{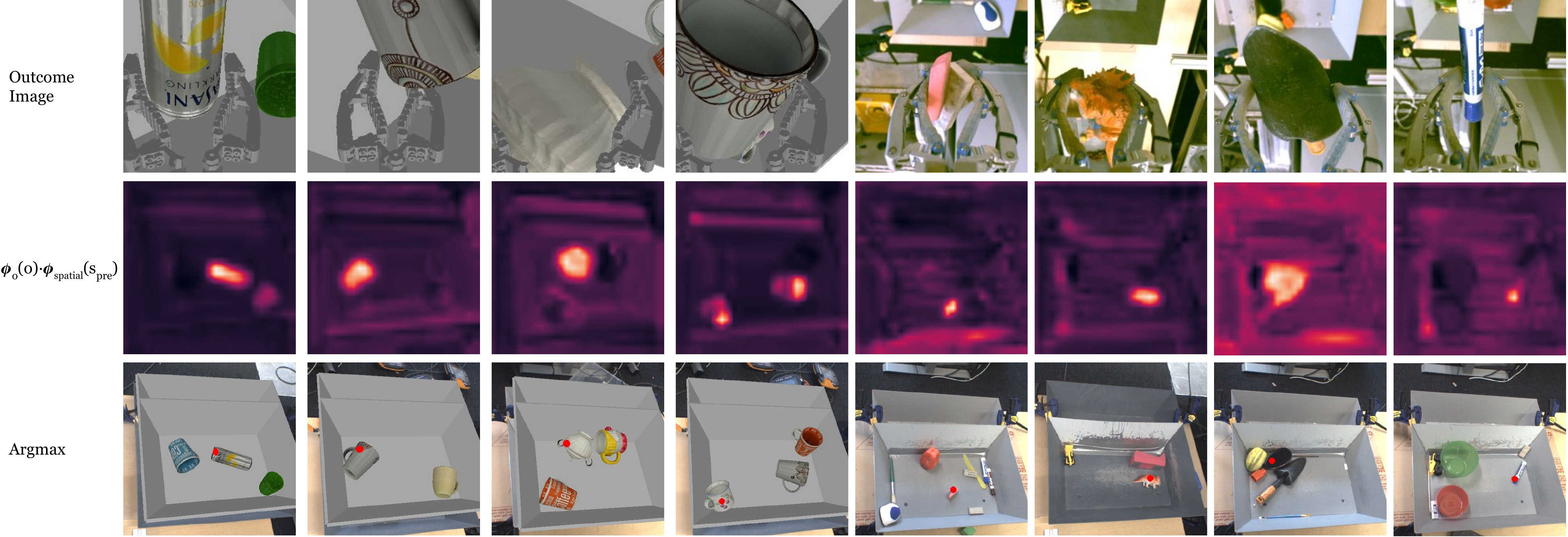}
}
\caption{An analysis of our learned embeddings. Examples shown were chosen at random from the dataset.}
\vspace{-0.2in}
\end{figure*}

\section{Self-Supervised Goal-Conditioned Grasping}
\label{sec:method_grasp}

The Grasp2Vec representation can enable effective goal-conditioned grasping, where a robot must grasp an object matching a user-provided query. In this setup, the same grasping system can both collect data for training the representation and utilize this representation for fulfilling specified goals.
The grasping task is formulated as a Markov decision process (MDP), similar to the indiscriminate grasping system proposed by~\citet{qtopt2018}. The actions $\ba$ correspond to Cartesian gripper motion and gripper opening and closing commands, and the state $\bs$ includes the current image and a representation of the goal $\goal$. We aim to learn the function $\Qpi(\bs, \ba, \goal)$ under the following reward function: grasping the object specified by $\goal$ yields a terminal reward of $\reward=1$, and $\reward=0$ for all other time steps. The architecture of $\Qpi$ is shown in Figure~\ref{fig:grasp2vec}.

Learning this Q-function presents two challenges that are unique to self-supervised goal-conditioned grasping: we must find a way to train the policy when, in the early stages of learning, it is extremely unlikely to pick up the right object, and we must also extract the reward from the episodes without ground truth object labels.

 \begin{wrapfigure}{R}{0.55\textwidth}
 \vspace{-0.5cm}
    \begin{minipage}{0.55\textwidth}
       \begin{algorithm}[H]
{\small
        \caption{Goal-conditioned policy learning}
        \label{alg:instancelabeling}
        \begin{algorithmic}[1]
        \STATE Initialize goal set \(G\) with 10 present images
        \STATE Initialize $\Qpi$ and replay buffer \(\replaybuffer\)
        \WHILE{$\pi$ not converged}
            \STATE $g, g^\prime \gets$ sample(\(G\))
            \STATE \((\bs, \ba, \rewardany, \outcome) \gets\) ExecuteGrasp($\pi, \goal$)
            \STATE //~Posthoc Labeling (PL)
            \IF{$\rewardany = 1$}
                \STATE \(\replaybuffer \gets \replaybuffer \bigcup (\bs, \ba, [\outcome, 1])\) (outcome is successful goal)
            \ELSE
                \STATE \(\replaybuffer \gets \replaybuffer \bigcup (\bs, \ba, [\goal, 0])\) (desired goal is a failure)
            \ENDIF

            \STATE //~Embedding Similarity (ES)
            \IF{$\rewardany = 1$}
          \STATE \(\replaybuffer \gets \replaybuffer \bigcup (\bs, \ba, [\goal, \hat{\phig}(\outcome) \cdot \hat{\phig}(\goal)])\)
          \STATE //~Auxiliary Goal (AUX)
          \STATE \(\replaybuffer \gets \replaybuffer \bigcup (\bs, \ba, [\othergoal, \hat{\phig}(\outcome) \cdot \hat{\phig}(\othergoal)])\)
            \ENDIF


            \STATE \((\bs, \ba, [\goal, \reward], \bsp) \gets \) sample(\(\replaybuffer\))
            \STATE \(\pi \gets \pi - \alpha \nabla_\pi (\Qpi(\bs, \ba, \goal) - (\reward + \gamma V_\pi(\bsp, \goal)))^2 \)
        \ENDWHILE
        \end{algorithmic}
}
\end{algorithm}
    \end{minipage}
  \end{wrapfigure}

We assume that the grasping system can determine automatically whether it successfully grasped an object, but not  which object was grasped. For example, the robot could check whether the gripper is fully closed to determine if it is holding something. We will use $\rewardany$ to denote the indiscriminate reward function, which is 1 at the last time step if an object was grasped, and 0 otherwise. Q-learning can learn from any valid tuple of the form \((\bs, \ba, \reward, \goal)\), so we use the $\rewardany$ to generate these tuples without object labels. We utilize three distinct techniques to automatically augment the training data for Q-learning, making it practical to learn goal-conditioned grasping:

\vspace{-0.1in}
\paragraph{Embedding Similarity (ES)}
A general goal labeling system would label rewards based on a notion of similarity between what was commanded, $\goal$, and what was achieved, $\outcome$, approximating the true on-policy reward function. If the Grasp2Vec representations capture this similarity between objects, setting $\reward = \hat{\phig}(\goal) \cdot \hat{\phig}(\outcome)$ would enable policy learning for instance grasping.

\vspace{-0.1in}
\paragraph{Posthoc Labeling (PL)}
\label{methods:posthoc}
Embedding similarity will give close to correct rewards to the Q function, but if the policy never grasps the right object there
will be no signal to learn from. We use a data augmentation approach similar to the hindsight experience replay technique proposed by \citet{hindsight2017}.
If an episode grasps any object, we can treat $\outcome$ as a correct goal for that episode's states and actions, and add the transition \((\bs, \ba, \outcome, \reward=1)\) to the buffer. We refer to this as the posthoc label.

\vspace{-0.1in}
\paragraph{Auxiliary Goal Augmentation (Aux)}
We can augment the replay buffer even further by relabling transitions with unacheived goals. Instead of sampling
a single goal, we sample a pair of goals \( (\goal, \othergoal) \) from the goal set $G$ without replacement
If \(\rewardany==1\) after executing the policy conditioned on $\goal$, we add the transition \((\bs, \ba, \othergoal, \reward=\hat{\phig}(\othergoal) \cdot \hat{\phig}(\outcome))\) to the replay buffer. In baselines
that do not use embeddings, the reward is replaced with 0 under the assumption that $\othergoal$ is unlikely
to be the grasped object.

Pseudocode for the goal-reward relabeling schemes, along with the self-supervised instance grasping routine, are summarized in Algorithm~\ref{alg:instancelabeling}.

\section{Experiments}
\label{sec:experiments}

Our experiments answer the following questions. For any grasp triplet $(\scenepre, \outcome, \scenepost)$, does the vector $\phis(\scenepre) - \phis(\scenepost)$ indicate which object was grasped? Can $\phimap(\scenepre)$ be used to localize objects in a scene? Finally, we show that instance grasping policies can be trained by using distance between grasp2vec embeddings as the reward function. \newline
\url{https://sites.google.com/site/grasp2vec/}

\paragraph{Experimental setup and data collection.}
Real-world data collection for evaluating the representation on real images was conducted using KUKA LBR iiwa robots with 2-finger grippers, grasping objects with various visual attributes and varying sizes from a bin.
Monocular RGB observations are provided by an over-the-shoulder camera. Actions $\ba$ are parameterized by a Cartesian displacement vector, vertical wrist rotation, and binary commands to open and close the gripper.
The simulated environment is a model of the real environment simulated with Bullet~\cite{bullet}.
Figure~\ref{fig:kuka_setup} depicts the simulated and real-world setup, along with the image observations from the robot's camera.We train and evaluate grasp2vec embeddings on both the real and simulated environments across 3 tasks: object recall from scene differences, object localization within a scene, and use as a reward function for instance grasping.

\subsection{Grasp2Vec embedding analysis.}
We train the goal and scene embeddings on successful grasps. We train on 15k successful grasps for  the simulated results and 437k for the real world results.  The objective pushes embeddings of outcome images to be close to embedding difference of their respective scene images, and far from each other.
By representing scenes as the sum of their objects, we expect the scene embedding space to be structured by object presence and not by object location. This is validated by the nearest neighbors of scenes images shown in Figure~\ref{fig:neighbors}, where nearest neighbors contain the same same objects regardless of position or pose.

\begin{table}[h]

\begin{center}
\begin{tabular}{|l|c|c|c|c|}
\hline
 & sim seen & sim novel & real seen & real novel\\
\hline
Retrieval (ours) & 88\%& 64\% & 89\% & 88\% \\
Outcome Neighbor (ImageNet) & \textemdash& \textemdash& 23\%& 22\% \\
\hline \hline
Localization (ours) & 96\% & 77\% & 83\% & 81\%\\ \hline
Localization (ImageNet) & \textemdash& \textemdash& 18\%& 15\% \\

\hline
\end{tabular}

\end{center}
\caption{Quantitative study of Grasp2Vec embeddings. As we cannot expect weights trained on ImageNet to exhibit the retrieval property, we instead evaluate whether two nearest neighbor outcome images contain the same object.
For object localization, the ImageNet baseline is performed the same as the grasp2vec evaluation. See Figure~\ref{fig:detection} for examples heatmaps and Appendix~\ref{app:embeddings} for example retrievals. \label{table:embeddings}}
\end{table}

\paragraph{For any grasp triplet $(\scenepre, \outcome, \scenepost)$, does the vector $\phis(\scenepre) - \phis(\scenepost)$ indicate which object was grasped?} We can evaluate the scene and and outcome feature spaces
by verifying that the the difference in scene features $\phis(\scenepre) - \phis(\scenepost)$ are near the grasped object features $\phig(\outcome)$.
As many outcome images of the same object will have similar features, we define the \textit{retrieval} as correct if the nearest neighbor outcome image contains the same object as the one grasped from $\scenepre$.
Appendix~\ref{app:embeddings} shows example successes and failures. As shown in Table~\ref{table:embeddings}, retrieval accuracy is high for the training simulated objects and all real objects. Because the simulated data contained fewer unique objects that the real dataset, it is not surprising that the embeddings trained on real images generalized better.

\paragraph{ Can $\phimap(\scenepre)$ be used to localize objects in a scene?} The grasp2vec architecture and objective enables our method to localize  objects without any spatial supervision.

By embedding scenes and single objects (outcomes) into the same space, we can use the outcome embeddings to localize that object within a scene. As shown in Figure~\ref{fig:detection}, we compute the dot product of $\phig(o)$ with each pixel of $\phi_{s,spatial}(\scenepre)$ to obtain a heatmap over the image corresponding to the affinity between the query object and each pixel's receptive field. A localization is considered correct only if the point of maximum activation in the heatmap lies on the correct object. As shown in Table~\ref{table:embeddings}, grasp2vec embeddings perform localization at almost 80\% accuracy on objects that were never seen during training, without ever receiving any position labels. The simulated objects seen during training are localized at even higher accuracy. We expect that such a method could be used to provide goals for pick and place or pushing task where a particular object position is desired. For this localization evaluation, we compare grasp2vec embeddings against the same ResNet50-based architecture used in the embeddings, but trained on ImageNet~\cite{russakovsky2015imagenet}. This network is only able to localize the objects at 15\% accuracy, because the features of an object in the gripper are not necessary similar to the features of that same object in the bin.

\subsection{Simulated instance grasping}
While past work has addressed self-supervised indiscriminate grasping, we show that instance grasping can be learned with no additional supervision.
We perform ablation studies in simulation and analyze how choices in model architecture and goal-reward relabeling affect instance grasping performance and generalization to unseen test objects.

Overall instance grasping performance is reported in Table~\ref{table:grasping}. All models using our method (Grasp2Vec embedding similarity) in the reward function achieve at least 78\% instance grasping success on seen objects. Our experiments yield the following conclusions:

\textit{How well can a policy learn without any measure of object similarity?} Looking at experiments 1 and 3, we see that posthoc labeling performs more than twice as well as indiscriminate grasping, while requiring no additional information. However, the PL only experiment is far behind the upper bound of using the true labels in experiment 3. Adding in auxiliary goal supervision in experiment 4, where data is augmented by randomly sampling a different goal image and marking it as a failed trajectory, only worsens the performance.

\textit{Does any embedding of the goal image provide enough supervision?} The goal of the reward label is to indicate whether two goal images contain the same object, in order to reward the policy for grasping the correct object. We already found in Table~\ref{table:embeddings} that ImageNet weights failed at this task. In experiment 5, we find that an autoencoder trained to encode goal images fails to provide a good reward label, performing no better than a indiscriminate policy.

\textit{How close can grasp2vec embeddings get to the oracle performance?}
The oracle labels requires knowing the true identity of the objects in the goal an outcome images. The cosine similarity of the goal and outcome image's grasp2vec embeddings approximates this label much better than the autoencoder embeddings. Experiments 6 and 7 show that using grasp2vec similarity leads to performance on-par to the oracle on objects seen in training, and outperform the oracle on new objects that the policy was trained on. Unlike the oracle, grasp2vec similarity requires no object identity labels during either training or testing.

\textit{Should the grasp2vec embedding also be used to condition the policy?} In experiment 8, we condition the policy on the embedding of the goal image instead of on the image itself. This reduces performance only on the unseen objects, indicating that the embeddings may hurt generalization by a small margin.

\begin{table}[h]

\begin{center}
\begin{tabular}{|l|c|c|c|c|}
\hline
\# & Goal Conditioning & Reward Labels & Seen Objects & Unseen Objects\\
\hline
1 & Indiscriminate Grasping & N/A & 18.3 & 21.9 \\
2 & Raw Image CNN & Oracle Labels & 83.9 & 43.7 \\
\hline
3 & Raw Image CNN & PL & 50.4 & 41.1 \\
4 & Raw Image CNN & PL + Aux(0) & 22.1 & 19.0 \\
5 & Raw Image CNN & PL + ES (autoencoder) & 18.7 & 20.9 \\
\hline
6 (ours) & Raw Image CNN & PL + ES (grasp2vec) & \textbf{80.1} & 53.9  \\
7 (ours) & Raw Image CNN & PL + Aux + ES (grasp2vec) & 78.0 & \textbf{58.9} \\
8 (ours)& $\phi_o(g)$& PL + ES (grasp2vec)& 78.4 & 45.4 \\
\hline
\end{tabular}

\end{center}
\caption{Evaluation and ablation studies on a simulated instance grasping task, averaged over 700 trials. In simulation, the scene graph is accessed to evaluate ground-truth performance, but it is withheld from our learning algorithms. Performance is reported as percentage of grasps that picked up the user-specified object. Table reports early stopping scores for instance grasping on training objects and evaluation for the same checkpoint on test objects. Best numbers (for unsupervised approaches) in bold font.\label{table:grasping}}
\end{table}

\paragraph{Composite goals.}
The additive compositionality of Grasp2Vec embeddings enables users to freely manipulate embeddings interactively to enable rich behavior at test time, without the policy ever having been explicitly trained to handle such goals. Our results show that policies conditioned on $\phig$ can grasp one of two simultaneously commanded goal embeddings that are averaged together: in simulation the composite-goal conditioned policies obtains \textbf{51.9\%} and \textbf{42.9\%} success for seen and unseen objects, respectively. The policy, which was only trained on single goals, generalizes to composite goals due to the additive semantics of grasp2vec embeddings.

\begin{wrapfigure}{r}{0.45\textwidth}
\vspace{-0.05in}
\begin{center}
\includegraphics[width=0.45\textwidth]{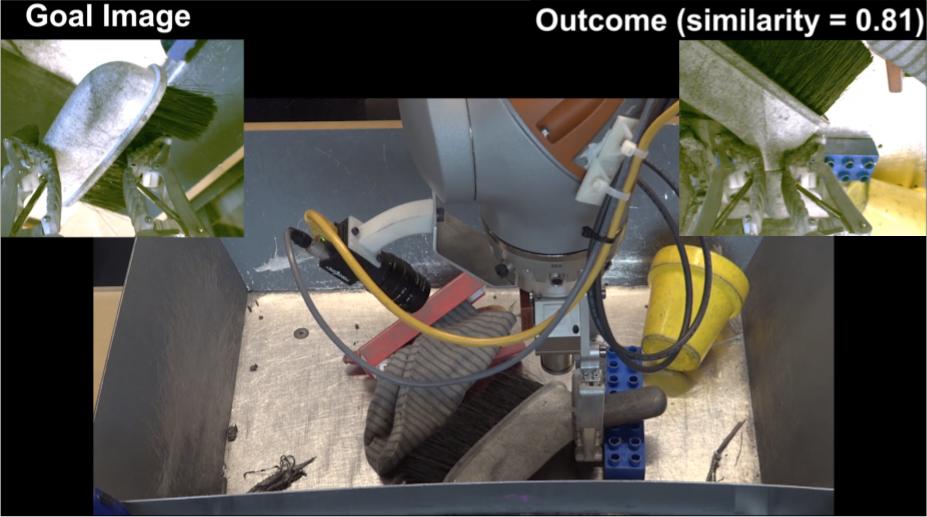}
\end{center}
\caption{Instance grasping with grasp2vec outcome similarity. The goal image is shown on the left, and the center shows the robot during the trajectory. The final outcome image is on the right along with its grasp2vec similarity to the goal.}
\label{fig:realgrasping}
\vspace{-0.2in}
\end{wrapfigure}

\subsection{Real-world instance grasping.}
To further evaluate the real-world grasp2vec embeddings, we design an instance grasping method that requires no additional on-policy training.
We leverage the localization property of grasp2vec embeddings and use a indiscriminate grasping policy (for example, the one used to collect the original grasping data). To grasp a goal object pictured in $\goal$ from a scene $s$, we obtain the 2D localization coordinate $(x,y) = \arg\max(\phig(\goal)*\phimap(s))$. Using the known camera calibration, we convert this into a 3D coordinate in the base frame of the arm and move the end-effector to that position. From there, an indiscriminate grasping policy is executed, which grasps the object closest to the gripper. To reduce the likelihood of accidentally picking up the wrong object, we compute the grasp2vec cosine similarity of a wrist-mounted camera image with the goal image. If the robot has just grasped an object and similarity falls below a threshold of 0.7, we drop the object and re-localize the goal object. We run this policy for a maximum of 40 time steps. Using this method, we obtain \textbf{80.8\%}  and \textbf{62.9\%} instance grasp success on training and test objects, respectively.

\section{Discussion}

We presented grasp2vec, a representation learning approach that learns to represent scenes as sets of objects, admitting basic manipulations such as removing and adding objects as arithmetic operations in the learned embedding space. Our method is supervised entirely with data that can be collected autonomously by a robot, and we demonstrate that the learned representation can be used to localize objects, recognize instances, and also supervise a goal-conditioned grasping method that can learn via goal relabeling to pick up user-commanded objects. Importantly, the same grasping system that collects data for our representation learning approach can also utilize it to become better at fulfilling grasping goals, resulting in an autonomous, self-improving visual representation learning method. Our work suggests a number of promising directions for future research: incorporating semantic information into the representation (e.g., object class), leveraging the learned representations for spatial, object-centric relational reasoning tasks (e.g.,~\cite{johnson2017clevr}), and further exploring the compositionality in the representation to enable planning compound skills in the embedding space.

\section{Acknowledgements}

We thank Ivonne Fajardo for running robot experiments, Julian Ibarz, Dmitry Kalashnikov, and Alex Irpan for the design of the QT-Opt system used for grasping evaluations, Ming Zhao, Ian Wilkes, Anthony Brohan, Peter Pastor, Adrian Li, Corey Lynch, Yunfei Bai, and Stephen James for helpful discussion and code reviews, and Tsung-Yi Lin, Chelsea Finn, and anonymous CoRL reviewers for helping proofread and improve earlier drafts of the paper.

\bibliography{example}
\newpage

\appendix

\section{Qualitative Detection Results}
\label{app:embeddings}
\begin{figure}[h]
\centering
\includegraphics[width=.8\textwidth]{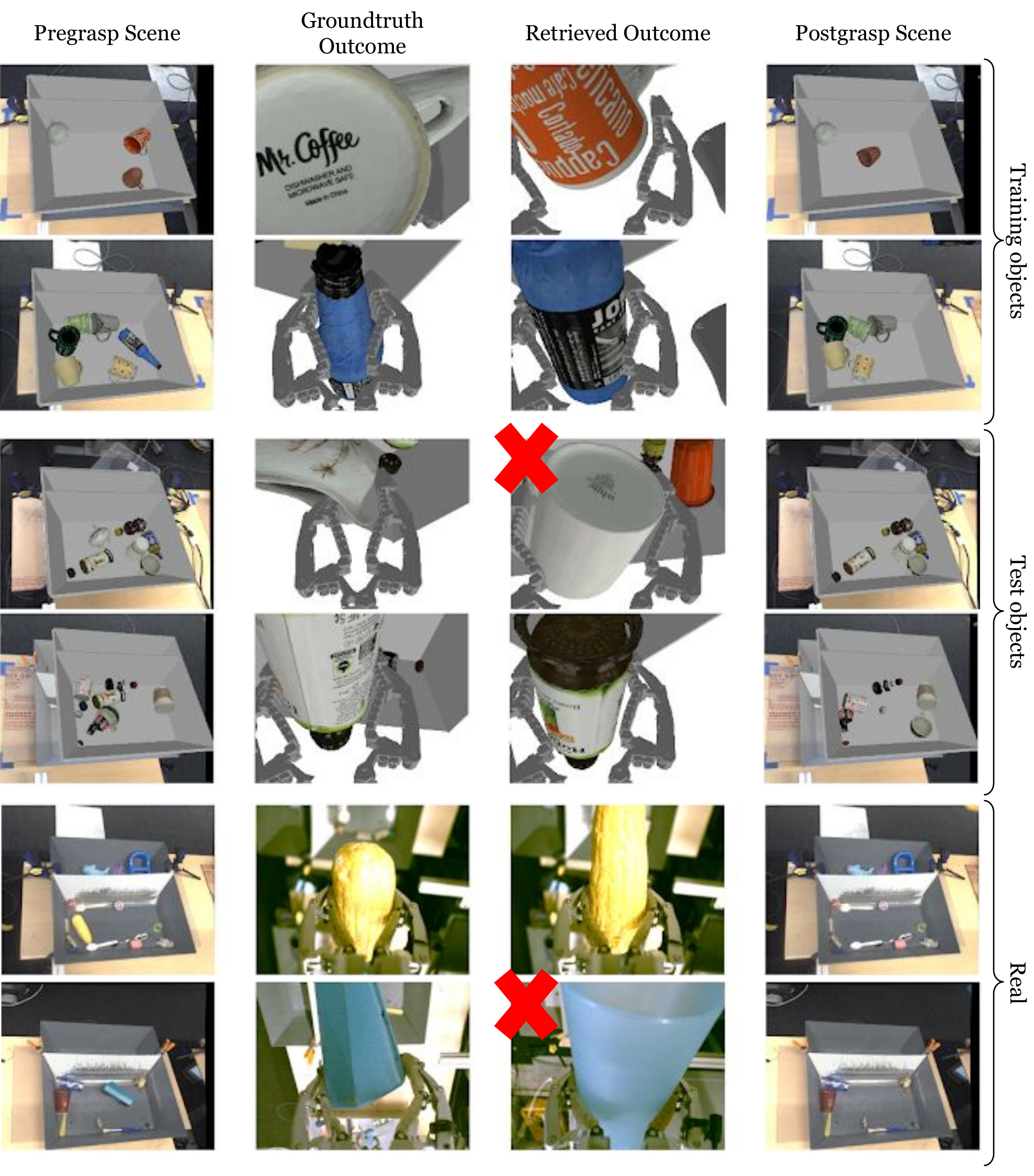}
\caption{This table illustrates the object recall property. From left to right: The scene before grasping, the grasped object, the outcome image retrieved by subtracting the postgrasp scene embedding from the pregrasp scene embedding, and lastly the postgrasp scene. We show example successes and failures (the later are marked with a red X). Failures occur in the test object set because multiple white objects had similar embeddings. Failures occur in the real data because the diversity of objects is very high, which likely makes the embddings less partitioned between object instances.}
\end{figure}

\begin{figure}[H]
\centering
\includegraphics[width=\textwidth]{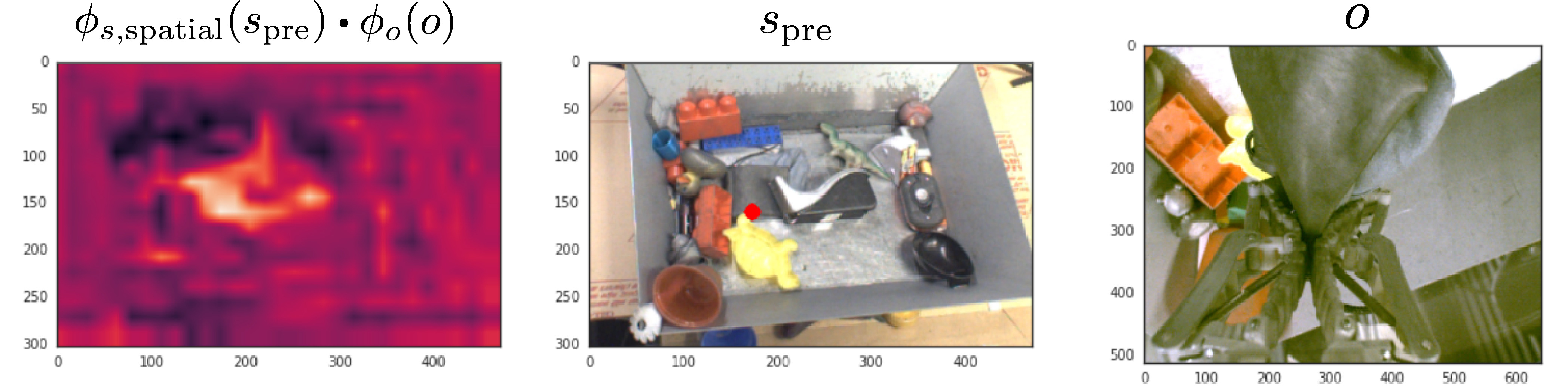}
\caption{Training objects. Recognize deformable object (bag) in a large amount of clutter.}
\label{fig:localize_deformable}
\end{figure}

\begin{figure}[H]
\centering
\includegraphics[width=\textwidth]{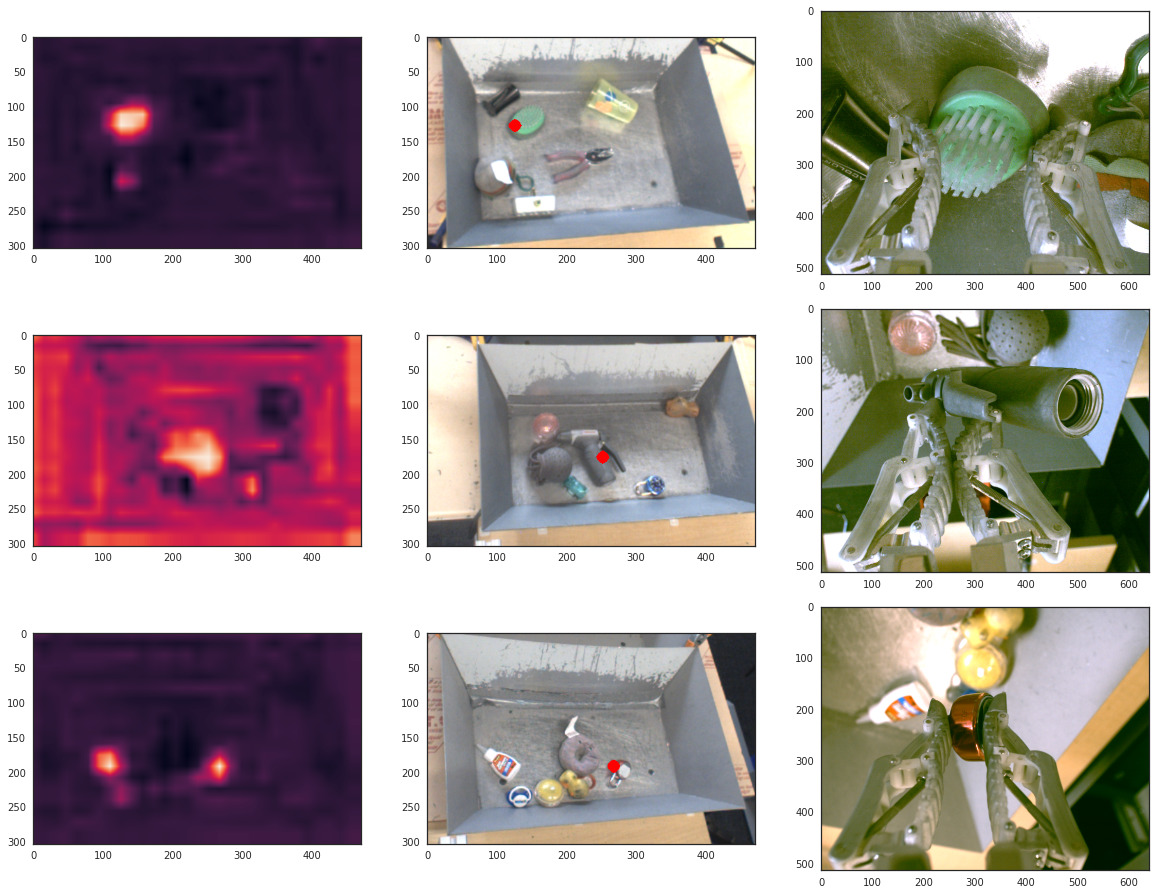}
\caption{Testing objects. Recognizes correct object, even when goal is presented from a different pose than the object's pose in the bin.}
\label{fig:localize_diff_pose}
\end{figure}

\begin{figure}[H]
\centering
\includegraphics[width=\textwidth]{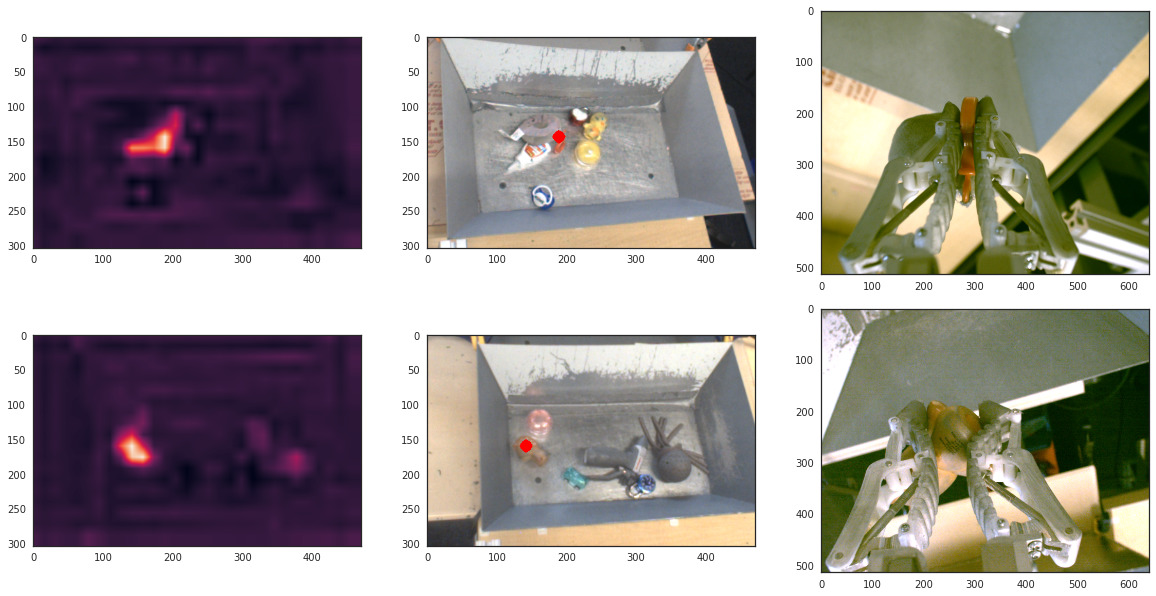}
\caption{Testing objects. Recognizes objects by parts when the goal object is occluded by the gripper.}
\label{fig:localize_parts}
\end{figure}

\paragraph{Success Cases}
Figures~\ref{fig:localize_deformable}, \ref{fig:localize_diff_pose}, \ref{fig:localize_parts} depict examples from the real-world localization task in which grasp2vec demonstrates surprisingly effective localization capabilities.

\begin{figure}[H]
\centering
\includegraphics[width=\textwidth]{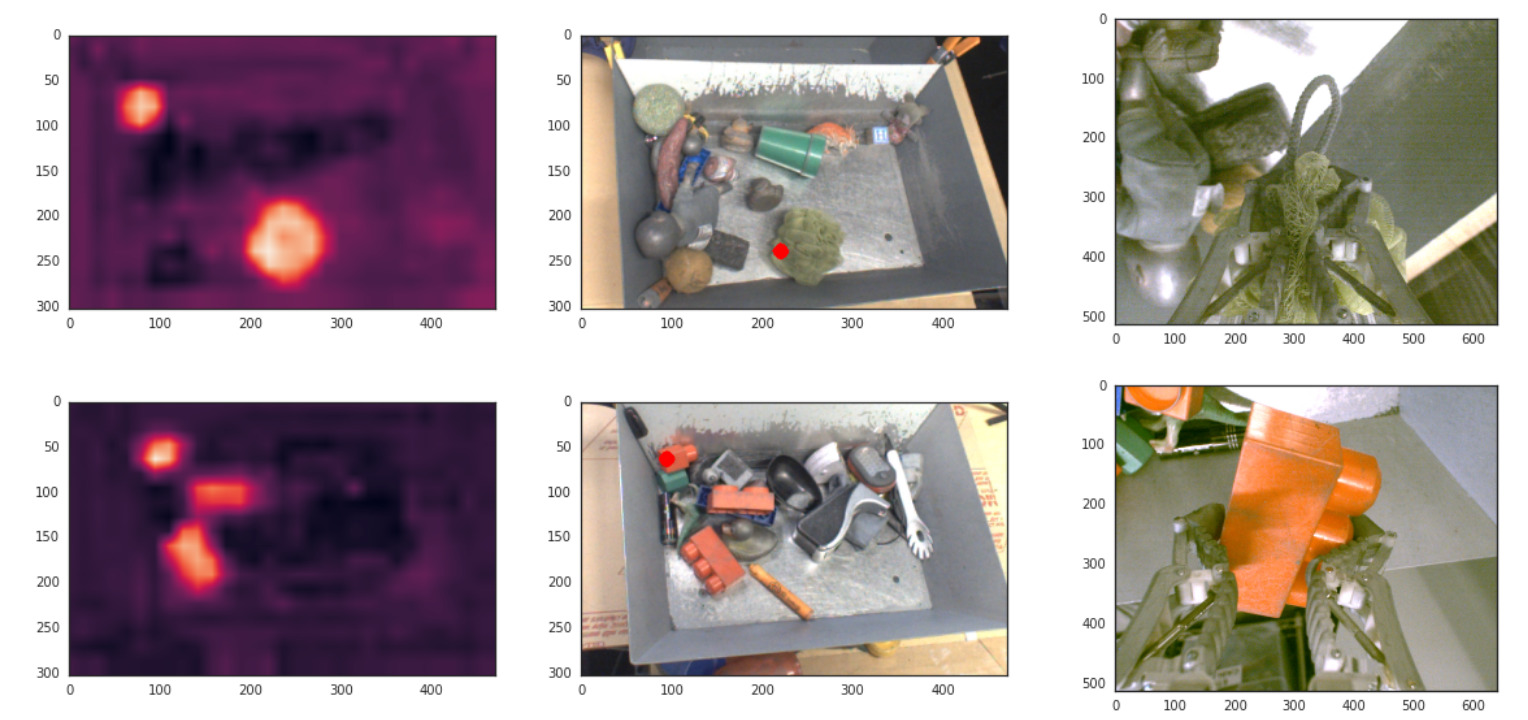}
\caption{Training objects. Embedding localizes objects with the correct colors but wrong (though similar) shape.}
\label{fig:localize_fail}
\end{figure}

\begin{figure}[H]
\centering
\includegraphics[width=\textwidth]{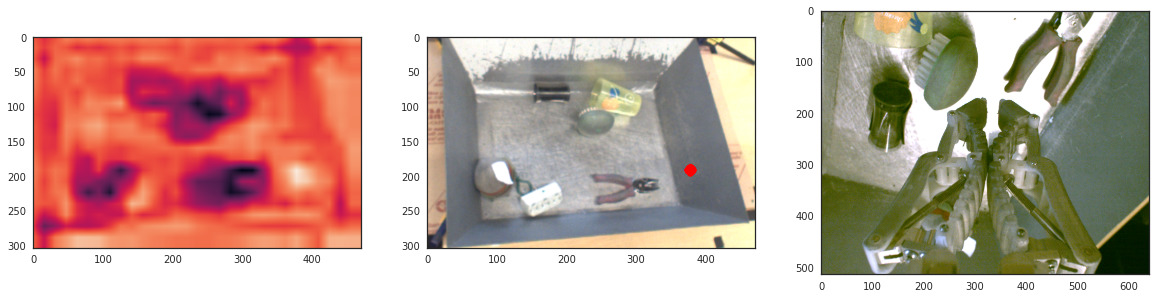}
\caption{Test objects. Failed grasp incorrectly labeled as successful, resulting in an empty (meaningless) goal for localization.}
\label{fig:localize_misgrasp}
\end{figure}

\paragraph{Failure Cases}
Figures~\ref{fig:localize_fail}, \ref{fig:localize_misgrasp} depict examples where Grasp2Vec embeddings make mistakes in localization.

\begin{figure}[h]
\centering
\subfloat[Localization using untrained weights in simulation.\label{fig:untrained}]{
\includegraphics[width=0.5\textwidth]{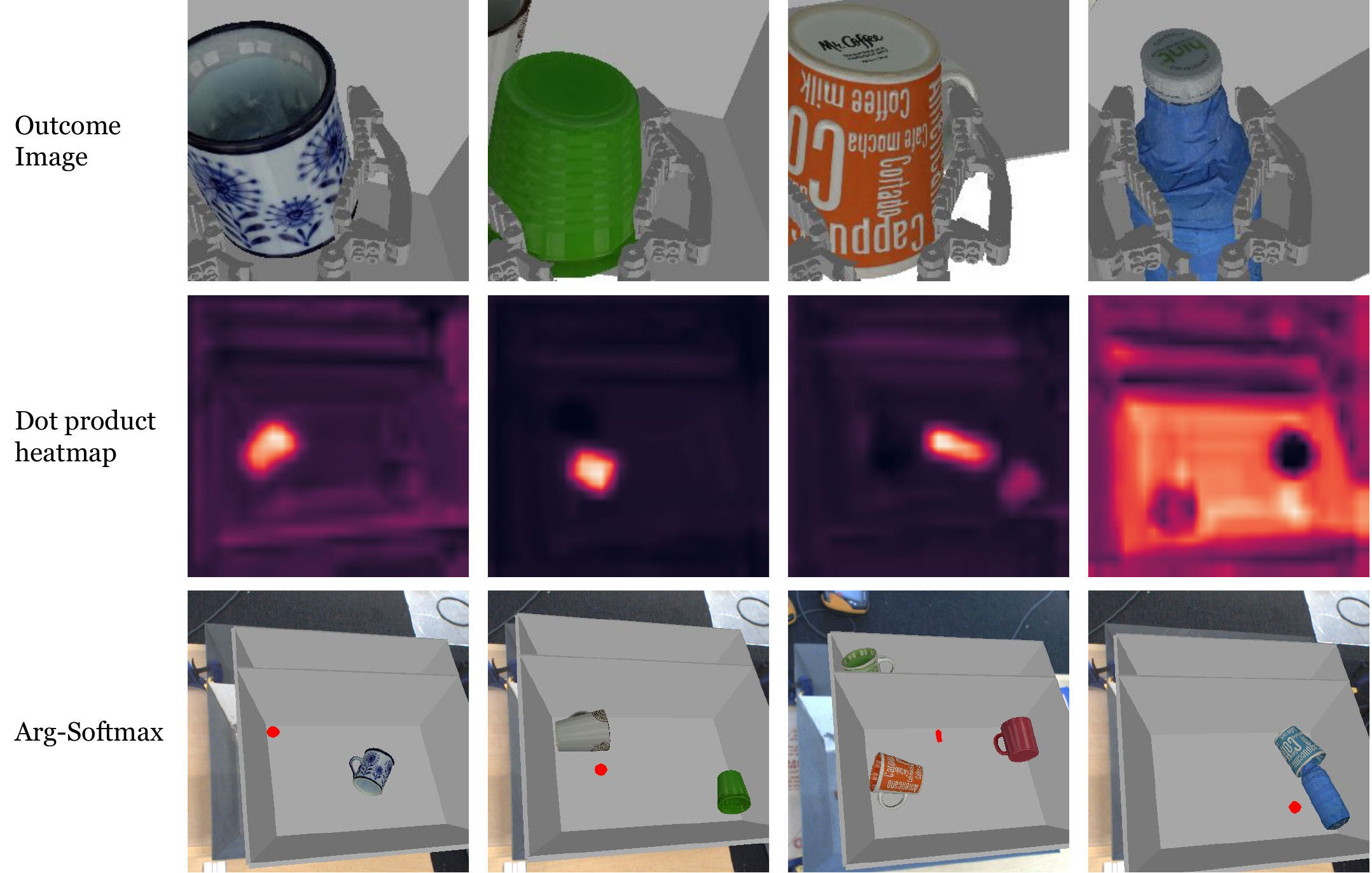}}
\subfloat[Localization using weights trained on imagenet.\label{fig:imagenet}]{
\includegraphics[width=0.5\textwidth]{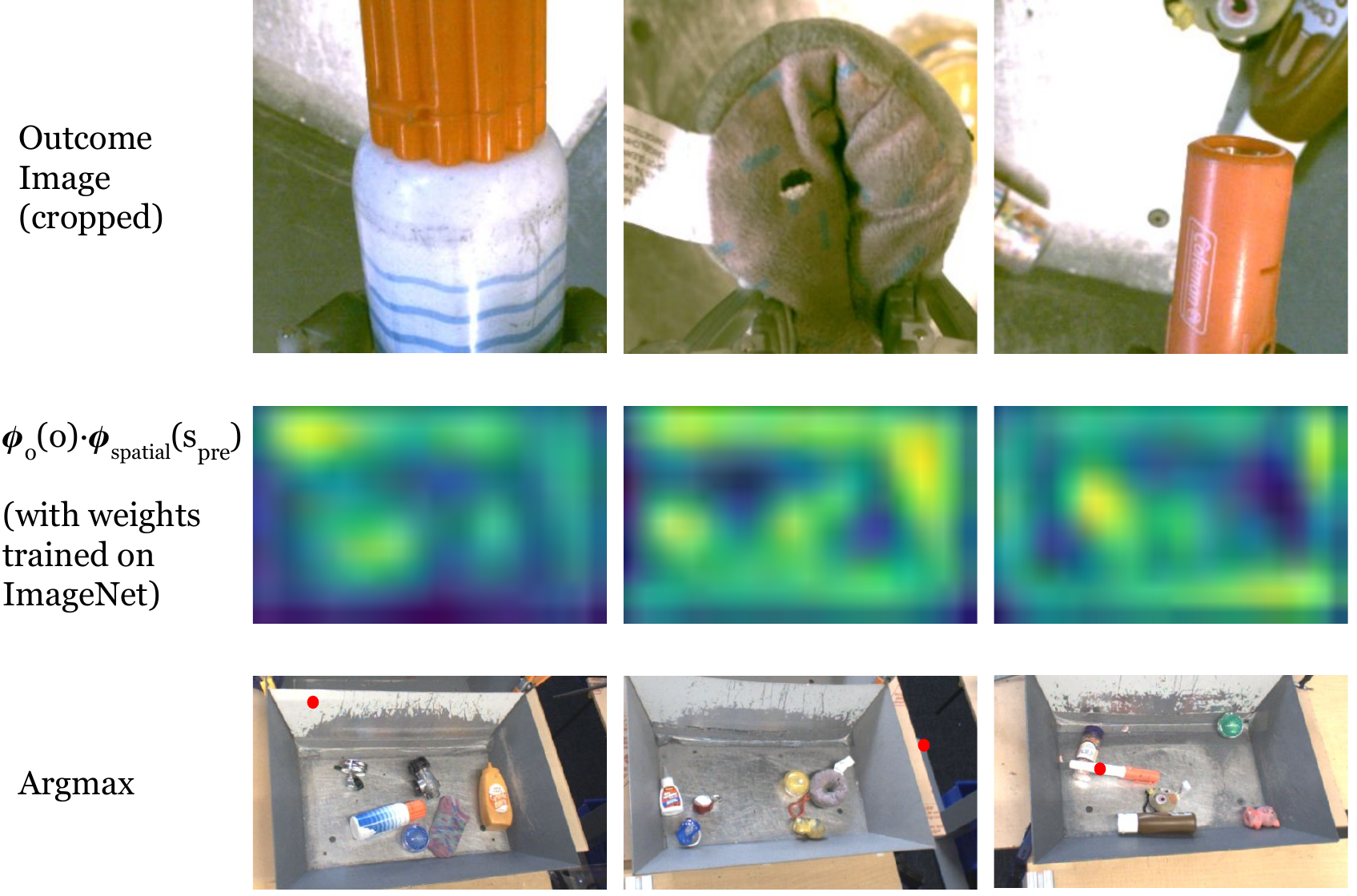}}

\caption{(a)The detection analysis with an untrained model. This verifies that our loss, rather than the architecture on it's own, enables the detection property. (b) The failure of localization indicates that Imagenet features are not consistent between scene and outcome images, probably because of resolution and pose differences.}
\end{figure}

\section{Simulation Experiment Details}
The simulated experiments use the Bullet~\cite{bullet} simulator with a model of a 7-DoF Kuka arm. We use 44 unique objects for training and 15 unique objects for evaluation. The training and evaluation objects are mutually exclusive, and each object has a unique mesh and texture. All objects are scans of mugs, bottles, cups, and bowls.

For data collection and evaluation, a particular scene has up to 6 objects sampled from the total objects without replacements; this means that no scene has multiple copies of a particular object. The objects are dropped into the scene at a random pose, and the objects may bounce onto each other.

To train the grasping policy, we use the following procedure. In each episode of the data collection and learning loop, an indiscriminate grasping policy collects 10 objects from the bin and saves their images to a goal set \(G\). Afterwards, the data collection protocol is switched to on-policy instance grasping,
using previously grasped images $\outcome^{(1)}...\outcome^{(10)}$ as subsequent goals. Exploration policy hyperparameters are as described in \cite{qtopt2018}, and we parallelize data collection across 1000 simulated robots for each experiment.

\section{Real-World Experiment Details}
We use roughly 500 objects for training and 42 unseen objects for evaluation. Objects are not restricted to object categories. For data collection and evaluation, 6 objects are placed randomly into the bin. After grasping an object, the robot drops it back into the bin to continue. The objects in a bin are switched out about twice a day and 6-7 robots are used in parallel, each with its own bin.

\begin{figure}[h]
\centering
\includegraphics[width=.8\textwidth]{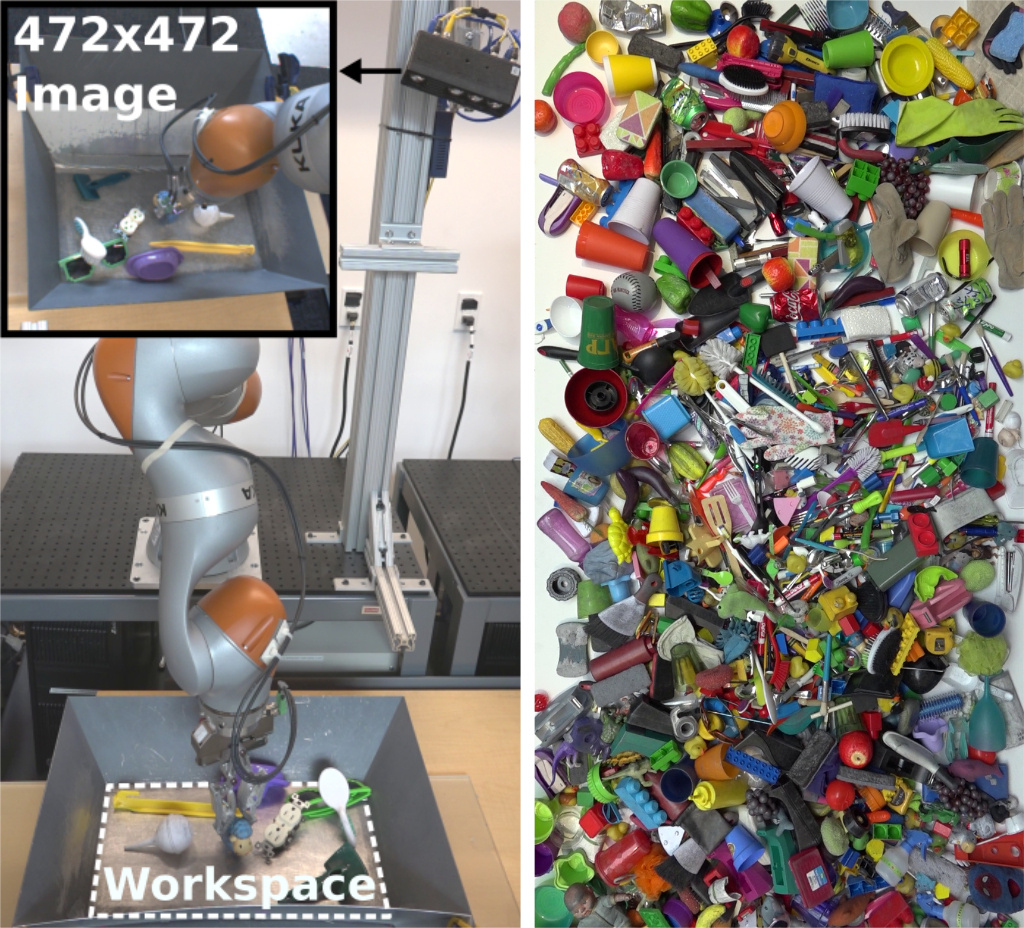}
\label{fig:kuka_real_setup}
\end{figure}

\begin{figure}[h]
\centering
\includegraphics[width=\textwidth]{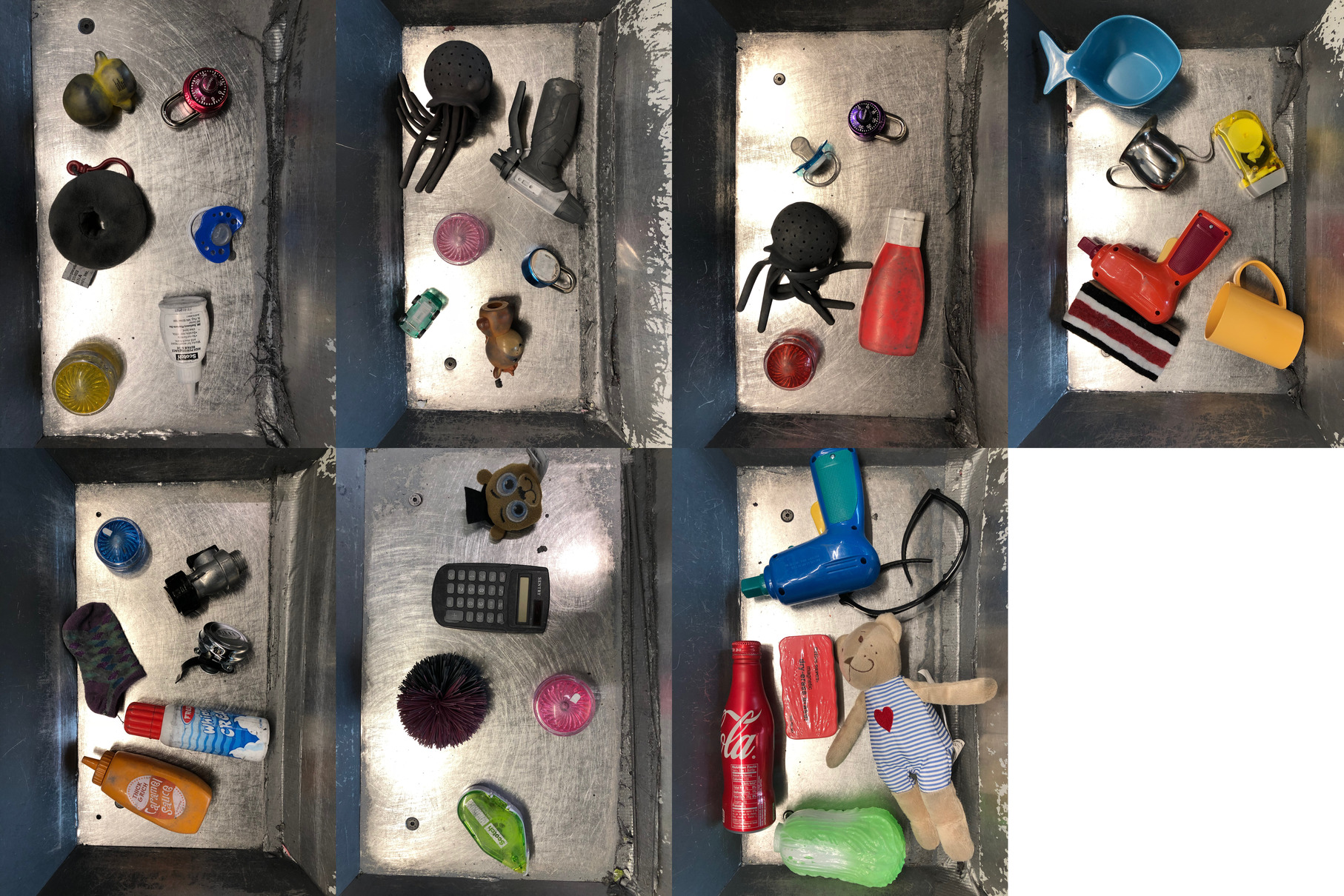}
\caption{Objects used for evaluation on unseen (test) objects.}
\label{fig:kuka_test_objects}
\end{figure}

\end{document}